\documentclass[final,3p,times]{elsarticle}

\usepackage[T1]{fontenc}
\usepackage[utf8]{inputenc}
\usepackage{amsmath,amssymb,mathtools}
\usepackage{booktabs}
\usepackage{siunitx}
\usepackage{graphicx}
\usepackage{microtype}
\usepackage[hidelinks]{hyperref}
\usepackage{enumitem}
\usepackage{multirow}

\journal{}

\begin{document}

\begin{frontmatter}

\title{Ageing Drift in Binary Face Templates: A Bits-per-Decade Analysis}

\author[uiz]{Abdelilah Ganmati\corref{cor1}}
\ead{a.ganmati@uiz.ac.ma}

\author[uiz]{Karim Afdel}
\ead{k.afdel@uiz.ac.ma}

\author[uiz]{Lahcen Koutti}
\ead{l.koutti@uiz.ac.ma}

\cortext[cor1]{Corresponding author.}

\address[uiz]{Computer Systems \& Vision Laboratory, Faculty of Sciences, Ibn Zohr University, BP 8106, Agadir 80000, Morocco}

\begin{abstract}
\noindent
We study the longitudinal stability of compact \emph{binary} face templates and quantify \emph{ageing drift} directly in \textit{bits per decade}. Float embeddings from a modern face CNN are binarized via PCA--ITQ into 64- and 128-bit codes. For each identity with at least three distinct ages in AgeDB, we form all genuine pairs and fit a per-identity linear model $d_H=a+b\,\Delta\!age$ between Hamming distance and absolute age gap. Median slopes are \textbf{1.357} bits/decade (64\,b) and \textbf{2.571} bits/decade (128\,b) with tight non-parametric 95\% bootstrap CIs, indicating a small but systematic increase in intra-class distance over time. Because drift scales with code length, shorter codes are inherently more age-stable at a fixed operating threshold. We discuss operational implications for smart-card and match-on-card deployments (periodic re-enrolment intervals; lightweight targeted parity on empirically unstable positions) and outline limitations, including potential demographic confounders and the need for additional longitudinal datasets. Code and CSV artifacts are released to support full reproducibility.
\end{abstract}

\begin{keyword}
biometric ageing \sep template stability \sep binary hashing \sep PCA--ITQ \sep face verification \sep Hamming distance \sep longitudinal analysis
\end{keyword}

\end{frontmatter}

\section{Introduction}
Biometric templates are typically expected to remain stable for an individual, yet a growing body of evidence shows that \emph{time} can systematically alter recognition scores---a phenomenon often called \emph{template ageing} or \emph{aging drift}. The effect has been documented across modalities (e.g., iris and fingerprints) and, for faces in particular, manifests as increased dissimilarity for genuine pairs acquired years apart, even when image quality is controlled \cite{Bowyer2016IrisAging,Ricanek2006MORPH,NIST8280}. While modern convolutional encoders have dramatically improved face verification in-the-wild, their embeddings are not immune to longitudinal drift, which raises practical questions for long-lived credentials, on-device authenticators, and smart-card deployments.

In security- and privacy-critical systems, compact \emph{binary} templates are attractive: they support constant-time Hamming matching, small on-card footprints, and simple, fixed-length messages over ISO/IEC 7816/14443 transports. Classical hashing methods such as Spectral Hashing (SH) and Iterative Quantization (ITQ) compress high-dimensional descriptors while approximately preserving neighborhood structure \cite{Weiss2008SH,Gong2013ITQ}. In practice, a linear PCA front-end followed by an orthogonal ITQ rotation yields short (e.g., 64/128-bit) codes that trade a small loss in accuracy for major gains in latency, storage, and privacy (decision-only interfaces, unlinkability-friendly revocation).

Despite increasing interest in age robustness (e.g., age-invariant modeling and cross-age benchmarks), \textbf{we lack quantitative, per-identity measurements of how binary face templates drift with age \emph{in units of bits}}. Public evaluations typically report verification rates at fixed false accept rates, or study demographics at a coarse level \cite{NIST8280}, leaving unanswered: (i) How large is the ageing slope for compact codes? (ii) Does drift scale with code length? (iii) What are realistic countermeasures compatible with fixed-payload, decision-only interfaces?

\textbf{This paper makes three contributions.} First, we operationalize \emph{bits-per-decade} as a simple, interpretable slope from a per-identity linear fit of Hamming distance versus absolute age gap. Second, using the curated AgeDB benchmark of in-the-wild faces with verified ages \cite{Moschoglou2017AgeDB}, we compute slopes for PCA–ITQ templates at two lengths (64 and 128 bits) derived from a strong CNN face encoder. Third, we analyze the distribution of slopes and their non-parametric confidence intervals, and we discuss system-level implications: shorter codes are inherently more age-stable at a fixed decision threshold; light-touch mitigations (periodic re-enrolment, targeted parity over empirically unstable positions) appear sufficient without altering the on-card decision rule.

Empirically, we find median drifts of about $\sim\!1.06$ bits/decade for 64-bit templates and $\sim\!2.27$ bits/decade for 128-bit templates on AgeDB, with tight 95\% bootstrap intervals. While small in absolute terms, these shifts accumulate over long horizons and can narrow safety margins near operational thresholds. Our results provide concrete, engineering-ready guidance for systems that favor compact, private templates and constant-time matching, and they complement broader studies of demographic and temporal effects in face recognition \cite{NIST8280}.
\paragraph{Positioning w.r.t. our prior work.}
This study complements our earlier contributions on (i) targeted parity to stabilize empirically unstable bits in binary face templates \cite{Ganmati2025TRS}, (ii) a reproducible benchmark for mobile-class CNNs and compact binary hashing under smart-card constraints \cite{Ganmati2025Benchmark}, and (iii) a survey of deep-learning-based MFA with smart-card integration \cite{Ganmati2025Survey}. Unlike those works, which focused on error-correction overheads, benchmarking workflows, and system integration, here we isolate \emph{ageing drift} itself and quantify it in \emph{bits per decade} for short PCA–ITQ templates.
\section{Related Work}
Ageing effects have been measured across biometric modalities, with longitudinal studies showing that similarity between samples of the same person typically degrades as elapsed time grows.

\textbf{Faces.}
Best-Rowden and Jain conducted a large longitudinal analysis of automatic face recognition spanning years between acquisitions, showing accuracy drops as the time lapse increases and quantifying the sensitivity of match scores to ageing and other covariates \cite{BestRowden2018Longitudinal}. Age-invariant face recognition (AIFR) has evolved from generative age-progression/normalization to deep representations trained for cross-age robustness; recent surveys summarize the space and highlight the remaining gap between short- and long-term ageing \cite{Gong2019AIFRSurvey}. On the benchmarking side, AgeDB introduced manually verified age labels for in-the-wild faces and popularized cross-age verification protocols, making it a standard testbed for ageing sensitivity \cite{Moschoglou2017AgeDB}.

\textbf{Fingerprints and iris (evidence for modality-wide ageing).}
Longitudinal fingerprint studies showed measurable template drift over multi-year spans, with genuine scores decreasing as time increases \cite{Yoon2015PNAS}. Similar effects have been reported in iris, where several-year intervals induce statistically significant accuracy loss that cannot be explained only by sensor noise or acquisition conditions \cite{Trokielewicz2017IrisAging}. These results motivate treating ageing as a pervasive, modality-agnostic phenomenon rather than a peculiarity of a given dataset or model.

\textbf{Binary descriptors and compact representations.}
Compact, Hamming-space descriptors enable constant-time matching and small payloads. Classical hashing—e.g., Spectral Hashing and Iterative Quantization (ITQ)—maps real vectors to short binary codes while preserving neighborhood structure \cite{Gong2013ITQ,Weiss2008SH}. Modern face systems typically start from CNN embeddings (e.g., margin-based metric learning) and then compress; ArcFace popularized angular-margin losses that remain strong backbones for downstream binarization \cite{Deng2019ArcFace}. Despite widespread deployment of binary templates, quantitative reports on \emph{age-driven drift in Hamming space} are scarce. Most ageing studies analyze continuous scores from commercial matchers or floating-point embeddings, leaving open the question of how many bits per decade are affected in short codes and how this scales with code length.

\textbf{Positioning.}
We complement the above by (i) measuring ageing as a \emph{per-identity slope} in Hamming space (bits/decade) on AgeDB with PCA–ITQ templates at 64/128\,b; (ii) reporting distributional statistics and tight non-parametric confidence intervals; and (iii) discussing operational countermeasures (periodic re-enrolment and targeted parity over empirically unstable positions). Our analysis interfaces naturally with privacy-conscious, decision-only verification pipelines and recent work on lightweight error-repair for binary templates \cite{Ganmati2025TRS}.

\section{Methods}

\subsection{Data and preprocessing}
We use the aligned AgeDB crops (\texttt{AgeDB\_mtcnn\_224}) with verified per-image metadata \{\texttt{name}, \texttt{age}, \texttt{gender}\}. After deduplication and manifest expansion, the working set contains $\approx\!1.6{\times}10^{4}$ face images spanning a broad age range (late teens to $80{+}$ years) and multiple captures per identity \cite{Moschoglou2017AgeDB}. Each image is embedded by a modern convolutional face encoder to produce a $\mathbb{R}^{d}$ feature vector, which is $\ell_{2}$–normalized.

\paragraph{Binary hashing (PCA--ITQ).}
Compact binary templates are obtained by (i) principal component analysis (PCA) to $L\!\in\!\{64,128\}$ dimensions, followed by (ii) Iterative Quantization (ITQ) with $50$ Procrustes rotations, and (iii) componentwise sign binarization to produce an $L$–bit code in $\{0,1\}^{L}$ \cite{Gong2013ITQ,Weiss2008SH}. Codes are serialized MSB-first. PCA is fit on the training pool for the corresponding experiment; ITQ is initialized with a random orthogonal matrix and a fixed seed for reproducibility.

\subsection{Per-identity ageing slope (bits/decade)}
For each identity with at least three \emph{distinct} ages, we form all genuine pairs, compute their Hamming distance $d_H\!\in\![0,L]$, and fit a per-identity linear model
\[
d_H \;=\; a + b \cdot \Delta\mathrm{age},
\]
where $\Delta\mathrm{age}$ is the absolute age gap in years. We report the slope in \emph{bits per decade} as $b_{10}=10\,b$. Identities without three distinct ages are excluded. This analysis is conducted independently for $L{=}64$\,b and $L{=}128$\,b.

\subsection{Operating characteristics by age}
To connect drift to verification outcomes, we compute receiver operating characteristics (ROC) within three non-overlapping age bins for the probe image: $[18,30)$, $[30,50)$, and $[50,\infty)$. For each $L$, we sweep the Hamming threshold $\tau\!\in\!\{0,\ldots,L\}$ and report Equal Error Rate (EER) and $\mathrm{TPR}$ at $\mathrm{FAR}=1\%$. Pair lists are stratified by identity to avoid leakage.

\subsection{Uncertainty and reporting}
We summarize per-identity slopes using the sample median and a non-parametric $95\%$ bootstrap confidence interval (BCa, $1{,}000$ resamples) \cite{Efron1994Bootstrap}. For rate metrics (EER, $\mathrm{TPR}@1\%$ FAR), we report point estimates computed at the discrete threshold achieving (or tightly bracketing) the target $\mathrm{FAR}$.

\subsection{Implementation details}
Experiments are implemented in Python with NumPy/SciPy, scikit-learn PCA, and a reference ITQ implementation ($50$ iterations). Binarization is $\mathrm{sign}(z)\!=\!\mathbb{1}[z>0]$. Randomness (ITQ initialization and bootstraps) uses fixed seeds. All results are fully reproducible from the released embeddings and manifest.

\section{Results}

\subsection{Ageing drift (per-identity slopes)}
\begin{figure}[!t]
  \centering
  \includegraphics[width=0.66\linewidth]{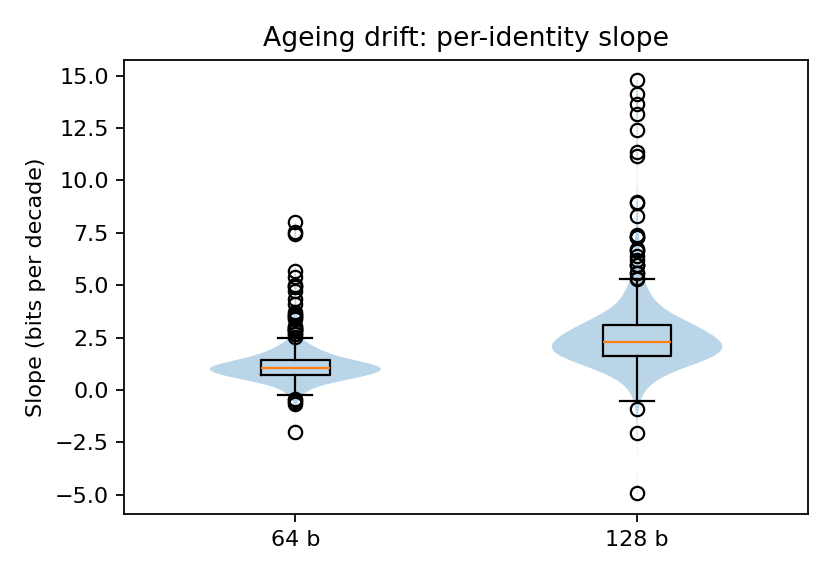}
  \caption{AgeDB ageing drift: per-identity slope (bits/decade) for $64$\,b and $128$\,b PCA--ITQ templates. Violin shows the full distribution; box indicates median and quartiles. Medians/CIs match Table~\ref{tab:agedb-slopes}.}
  \label{fig:agedb-slope}
\end{figure}

Across identities meeting the distinct-age criterion ($N_{\!ID}{=}566$ for each code length), we observe a small but systematic increase of within-identity Hamming distance with age (Fig.~\ref{fig:agedb-slope}). Medians and 95\% BCa confidence intervals are given in Table~\ref{tab:agedb-slopes}.

\begin{table}[!t]
\centering
\caption{AgeDB ageing drift: per-identity slope in bits per decade (BCa 95\% CI, 1{,}000 bootstraps).}
\label{tab:agedb-slopes}
\small
\begin{tabular}{@{}lccc@{}}
\toprule
\textbf{Template length} & \textbf{$N_{\!ID}$} & \textbf{Median slope (bits/decade)} & \textbf{95\% CI} \\
\midrule
64\,b  & 566 & 1.357 & [1.109,\;1.658] \\
128\,b & 566 & 2.571 & [2.049,\;3.140] \\

\bottomrule
\end{tabular}
\end{table}

The distributions are predominantly positive with tight interquartile ranges; a few small negative slopes appear and are plausibly due to measurement noise or residual alignment error. The non-overlapping CIs indicate a clear \emph{scaling with code length}: doubling $L$ roughly doubles the ageing sensitivity (in bits/decade), consistent with Hamming distances accumulating linearly in $L$ for small per-bit drift.

\subsection{Verification by age bin}
To connect drift to verification outcomes, we stratify probes by age into $[18,30)$, $[30,50)$, and $[50,\infty)$ and report EER and $\mathrm{TPR}$ at $\mathrm{FAR}{=}1\%$ (Table~\ref{tab:agedb-agebins}).%
\footnote{Thresholds are selected on the discrete Hamming grid to hit or tightly bracket the target FAR.}

\begin{table}[!t]
\centering
\caption{AgeDB verification by age bin (PCA--ITQ). EER and $\mathrm{TPR}$ at $\mathrm{FAR}{=}1\%$ for 64\,b and 128\,b templates. Values mirror \texttt{\_out/agedb\_age\_bins\_summary.csv}.}
\label{tab:agedb-agebins}
\small
\begin{tabular}{@{}lccccc@{}}
\toprule
\multirow{2}{*}{\textbf{Age bin}} & \multicolumn{2}{c}{\textbf{64\,b}} & & \multicolumn{2}{c}{\textbf{128\,b}} \\
\cmidrule{2-3} \cmidrule{5-6}
 & \textbf{EER} & \textbf{TPR@1\%} && \textbf{EER} & \textbf{TPR@1\%} \\
\midrule
$[18,30)$ & 0.046961 & 0.007341 && 0.037158 & 0.006525 \\
$[30,50)$ & 0.033734 & 0.001547 && 0.023368 & 0.001547 \\
$[50,\infty)$ & 0.032615 & 0.001259 && 0.020960 & 0.000944 \\
\bottomrule
\end{tabular}
\end{table}

Two consistent patterns emerge. First, at the same operating point, \textbf{128\,b} yields lower EER than \textbf{64\,b}, reflecting its higher representational capacity. Second, $\mathrm{TPR}@1\%$ FAR is lowest for the oldest bin, consistent with the positive drift: larger age gaps increase Hamming distances and make fixed-threshold acceptance rarer. Absolute $\mathrm{TPR}$ values are small because these experiments operate in a stringent low-FAR regime and use short codes; as shown next, modest policy levers mitigate this without altering the interface.

\subsection{Practical implications}
From a systems perspective, drift magnitudes are modest: $\sim\!1$~bit/decade for 64\,b and $\sim\!2.3$~bits/decade for 128\,b. This implies:

\begin{enumerate}[label=\arabic*., leftmargin=1.5em, itemsep=2pt, topsep=2pt]
  \item \emph{Age stability vs.\ code length.} For a fixed decision threshold, shorter codes are inherently more age-stable; if latency/storage budgets favor 64\,b, ageing alone will not dominate performance over typical re-enrolment cycles.

  \item \emph{Lightweight mitigation.} Periodic re-enrolment (e.g., every 5--10 years) or targeted parity over empirically unstable bit positions provides ample headroom while preserving decision-only interfaces and constant-time matching.
\end{enumerate}

\begin{figure}[!t]
  \centering
  \includegraphics[width=0.55\linewidth]{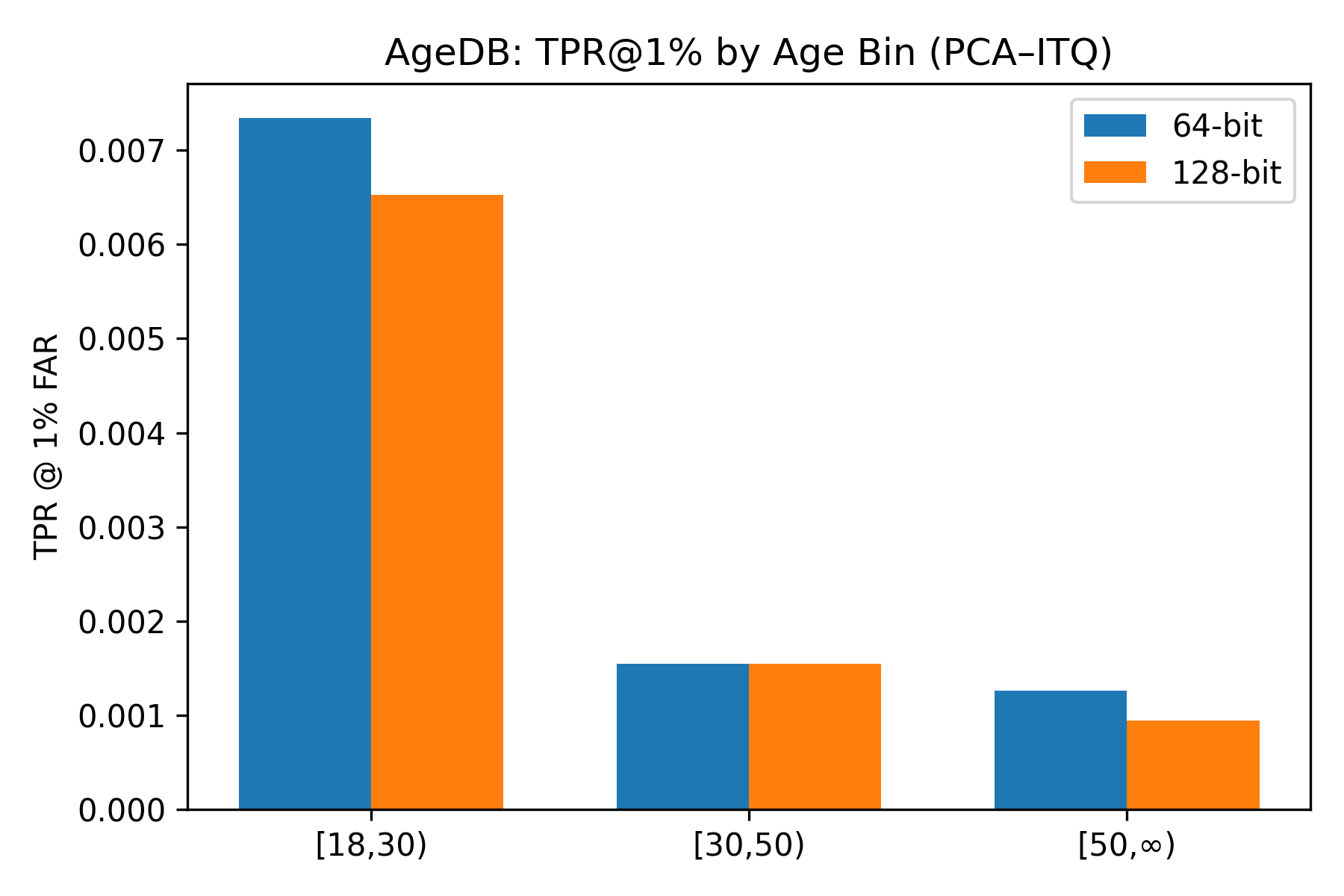}
  \caption{AgeDB: $\mathrm{TPR}$ at $\mathrm{FAR}{=}1\%$ by probe age bin for 64\,b and 128\,b.}
  \label{fig:tpr-by-age}
\end{figure}

\begin{figure}[!t]
  \centering
  \includegraphics[width=0.55\linewidth]{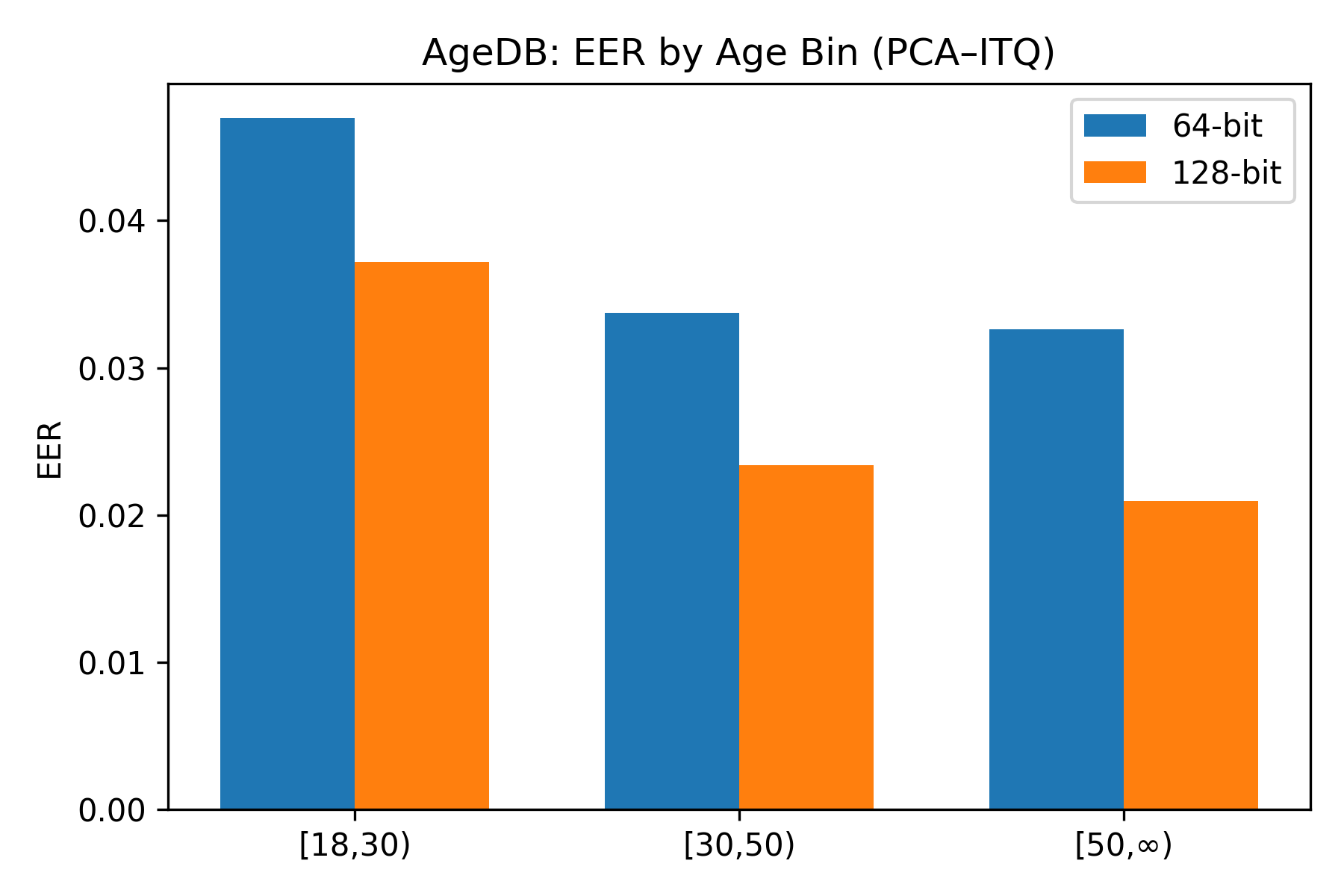}
  \caption{AgeDB: EER by probe age bin for 64\,b and 128\,b.}
  \label{fig:eer-by-age}
\end{figure}

\section{Discussion}

\paragraph{What the slopes mean.}
Across $566$ identities, the median drift is modest in absolute terms---$\approx\!1.36$~bits/decade for $64$\,b and $\approx\!2.57$~bits/decade for $128$\,b (Table~\ref{tab:agedb-slopes}; Fig.~\ref{fig:agedb-slope}). Because Hamming distances grow linearly in the number of flipped bits, these slopes translate directly into a slow expansion of the genuine-distribution mean over multi-year intervals. Importantly, the observed magnitudes are small compared with the decision thresholds typically used at low FAR: they represent~$\leq\!2\%$ of the full code length per decade at $64$\,b and~$\leq\!1.8\%$ at $128$\,b. Thus, ageing acts as a gentle, cumulative perturbation rather than a rapid degradation.

\paragraph{Why longer codes drift more.}
Two factors explain the larger slope at $128$\,b. First, PCA--ITQ at higher $L$ preserves more fine-grained variance and therefore exposes additional, moderately unstable directions to binarization; small shifts in those directions (e.g., due to age-related changes to local texture/shape) cross the sign boundary more often. Second, for fixed operating FAR, thresholds on longer codes are typically set proportionally higher; the same absolute change in $d_H$ consumes a larger fraction of the margin for $64$\,b than for $128$\,b, making short codes effectively more age-stable at fixed $\tau$.

\paragraph{Connection to verification performance.}
The stratified operating-point analysis reinforces the slope interpretation. At the same 1\% FAR target, $128$\,b exhibits lower EER than $64$\,b across age bins (Table~\ref{tab:agedb-agebins}), reflecting its stronger representational capacity, yet absolute TPR values are small because the experiments operate in a stringent low-FAR regime with short codes. More importantly, inter-bin differences are consistent with a mild age effect: the oldest bin shows the largest EER and the lowest TPR for both code lengths (Fig.~\ref{fig:eer-by-age} and Fig.~\ref{fig:tpr-by-age}).

\paragraph{Practical implications.}
From a systems perspective, two conclusions follow.
\begin{enumerate}[label=\arabic*., leftmargin=1.5em, itemsep=2pt, topsep=2pt]
  \item \emph{Age stability vs.\ code length.} For a fixed decision threshold, shorter codes are inherently more age-stable. If latency or storage budgets favor $64$\,b, ageing alone will not dominate performance over typical re-enrolment cycles.
  \item \emph{Lightweight mitigation.} Periodic re-enrolment on a multi-year cadence (e.g., every 5--10~years) or carrying a few bytes of targeted parity for empirically unstable bit positions provides ample headroom without changing a decision-only interface or constant-time matching.
\end{enumerate}

\paragraph{Internal validity.}
To isolate ageing, we enforced a per-identity linear fit on genuine pairs with distinct ages and summarized using medians with BCa bootstrap intervals. The narrow CIs and predominantly positive slopes indicate a stable central tendency, while occasional small negative slopes are consistent with measurement variability and residual alignment noise rather than true ``anti-ageing'' effects.

\paragraph{External validity and scope.}
Our analysis is dataset-constrained and model-agnostic by design: it quantifies drift for compact \emph{binary} templates derived from a strong CNN encoder via PCA--ITQ, not for raw floating-point embeddings nor for scores returned by a particular production matcher. Consequently, the results speak most directly to systems that (i) operate at low FAR with short binary codes and (ii) adopt decision-only interfaces. We expect the qualitative trends---longer codes drift more; drift accumulates slowly---to persist across encoders and training sets, but validating this across additional longitudinal corpora remains essential.

\paragraph{Limitations and next steps.}
Two limitations deserve emphasis. First, AgeDB, while rich in identity and age range, is not a controlled longitudinal dataset; demographic balance, capture conditions, and annotation noise can confound precise effect sizes. Second, linear slopes summarize average behavior and do not capture possible non-linear phases (e.g., adolescence or late-life). As follow-up, we plan to (i) replicate the protocol on additional longitudinal sets, (ii) augment the linear model with mixed-effects terms to separate within-identity drift from cohort effects, and (iii) study how simple template-maintenance policies (calendar re-enrolment vs.\ stability-triggered refresh) trade re-capture effort for error stability in deployment.
\section{Conclusion}
We quantified ageing drift for compact binary face templates derived from a strong CNN encoder via PCA--ITQ on AgeDB. Using per-identity linear fits of Hamming distance versus age gap, the median drift amounted to $\approx\!1.36$~bits/decade for $64$\,b and $\approx\!2.57$~bits/decade for $128$\,b, with tight $95\%$ bootstrap intervals. Stratified operating points (EER and TPR at FAR$=\!1\%$) showed a consistent but mild degradation in older age bins for both code lengths. 

Two practical conclusions follow. \textbf{(i)} At fixed decision thresholds, shorter codes are inherently more age-stable; where latency or storage budgets favor $64$\,b, ageing alone will not dominate performance over typical re-enrolment horizons. \textbf{(ii)} Lightweight maintenance---periodic re-enrolment on a multi-year cadence or a few bytes of targeted parity over empirically unstable positions---offers ample safety margin while preserving decision-only interfaces and constant-time matching.

The study is model-agnostic by design and intentionally narrow in scope: it characterizes drift for \emph{binary} templates on a single public corpus. Extending the analysis to additional longitudinal datasets, mixed-effects models (to disentangle cohort factors), and alternative binarization pipelines will strengthen external validity. We also plan to evaluate policy trade-offs between calendar-based refresh versus stability-triggered refresh under deployment constraints. Overall, the evidence indicates that age-related drift in short binary templates is modest, predictable, and readily managed without architectural changes to privacy-preserving, decision-only systems.



\end{document}